\documentclass[runningheads]{llncs}
\usepackage{graphicx}
\usepackage{subfig}
\usepackage{hyperref}
\usepackage{amsfonts}
\usepackage[subnum]{cases}
\usepackage{algorithm, algorithmic}

\begin{document}
\title{Queue-based Resampling for Online Class Imbalance Learning}
%
%
\author{Kleanthis Malialis \and Christos Panayiotou \and Marios M. Polycarpou}
\authorrunning{K. Malialis et al.}
%
\institute{
KIOS Research and Innovation Center of Excellence,\\Department of Electrical and Computer Engineering,\\University of Cyprus\\
\email{\{malialis.kleanthis, christosp, mpolycar\}@ucy.ac.cy}
}
\maketitle              
\begin{abstract}
Online class imbalance learning constitutes a new problem and an emerging research topic that focusses on the challenges of online learning under class imbalance and concept drift. Class imbalance deals with data streams that have very skewed distributions while concept drift deals with changes in the class imbalance status. Little work exists that addresses these challenges and in this paper we introduce queue-based resampling, a novel algorithm that successfully addresses the co-existence of class imbalance and concept drift. The central idea of the proposed resampling algorithm is to selectively include in the training set a subset of the examples that appeared in the past. Results on two popular benchmark datasets demonstrate the effectiveness of queue-based resampling over state-of-the-art methods in terms of learning speed and quality.

\end{abstract}
\section{Introduction}
In the area of monitoring and security of critical infrastructures which include large-scale, complex systems such as power and energy systems, water, transportation and telecommunication networks, the challenge of the state being normal or healthy for a sustained period of time until an abnormal event occurs is typically encountered \cite{kyriakides2014intelligent}. Such abnormal events or faults can lead to serious degradation in performance or, even worse, to cascading overall system failure and breakdown. The consequences are tremendous and may have a huge impact on everyday life and well-being. Examples include real-time prediction of hazardous events in environment monitoring systems and intrusion detection in computer networks. In critical infrastructure systems the system is at a healthy state the majority of the time and failures  are low probability events, therefore, class imbalance is a major challenge encountered in this area.

Class imbalance occurs when at least one data class is under-represented compared to others, thus constituting a minority class. It is a difficult problem as the skewed distribution makes a traditional learning algorithm ineffective, specifically, its prediction power is typically low for the minority class examples and its generalisation ability is poor \cite{wang2018systematic}. The problem becomes significantly harder when class imbalance co-exists with concept drift. There exists only a handful of work on online class imbalance learning. Focussing on binary classification problems, we introduce a novel algorithm, queue-based resampling, where its central idea is to selectively include in the training set a subset of the negative and positive examples by maintaining a separate queue for each class. Our study examines two popular benchmark datasets under various class imbalance rates with and without the presence of drift. Queue-based resampling outperforms state-of-the-art methods in terms of learning speed and quality.

\section{Background and Related Work}

\subsection{Online Learning}
In \textbf{online learning} \cite{ditzler2015learning}, a data generating process provides at each time step $t$ a sequence of examples $(x^t,y^t)$ from an unknown probability distribution $p^{t}(x,y)$, where $x^t \in \mathbb{R}^d$ is an $d$-dimensional input vector belonging to input space $X$ and $y^t \in Y$ is the class label where $Y=\{c_1, .., c_N\}$ and $N$ is the number of classes. An online classifier is built that receives a new example $x^t$ at time step $t$ and makes a prediction $\hat{y}^t$. Specifically, assume a concept $h: X \to Y$ such that $\hat{y}^t = h(x^t)$. The classifier after some time receives the true label $y^t$, its performance is evaluated using a loss function $J = l(y^t,\hat{y}^t)$ and is then trained i.e. its parameters are updated accordingly based on the loss $J$ incurred. The example is discarded to enable learning in high-speed data streaming applications. This process is repeated at each time step. Depending on the application, new examples do not necessarily arrive at regular and pre-defined intervals.

We distinguish \textbf{chunk-based learning} \cite{ditzler2015learning} from online learning where at each time step $t$ we receive a chunk of $M > 1$ examples $C^t = \{(x^t_i,y^t_i)\}^M_{i=1}$. Both approaches build a model incrementally, however, the design of chunk-based algorithms differs significantly and, therefore, the majority is typically not suitable for online learning tasks \cite{wang2018systematic}. This work focuses on online learning.

\subsection{Class Imbalance and Concept Drift}
\textbf{Class imbalance} \cite{he2008learning} constitutes a major challenge in learning and occurs when at least one data class is under-represented compared to others, thus constituting a minority class. Considering, for example, a binary classification problem, class $1$ (positive) and $0$ (negative) constitutes the minority and majority class respectively if $p(y=1) << p(y=0)$. Class imbalance has been extensively studied in offline learning and techniques addressing the problem are typically split into two categories, these are, data-level and algorithm-level techniques.

Data-level techniques consist of resampling techniques that alter the training set to deal with the skewed data distribution, specifically, oversampling techniques ``grow'' the minority class while undersampling techniques ``shrink'' the majority class. The simplest and most popular resampling techniques are random oversampling (or undersampling) where data examples are randomly added (or removed) respectively \cite{zhou2006training,wang2018systematic}. More sophisticated resampling techniques exist, for example, the use of Tomek links discards borderline examples while the SMOTE algorithm generates new minority class examples based on the similarities to the original ones. Interestingly, sophisticated techniques do not always outperform the simpler ones \cite{wang2018systematic}. Furthermore, since their mechanism relies on identifying relations between training data, it is difficult to be applied in online learning tasks, although some initial effort has been recently made \cite{mao2015online}.

Algorithm-level techniques modify the classification algorithm directly to deal with the imbalance problem. Cost-sensitive learning is widely adopted and assigns a different cost to each data class \cite{zhou2006training}. Alternatives are threshold-moving \cite{zhou2006training} methods where the classifier's threshold is modified such that it becomes harder to misclassify minority class examples. Contrary to resampling methods that are algorithm-agnostic, algorithm-level methods are not as widely used \cite{wang2018systematic}.

A challenge in online learning is that of \textbf{concept drift} \cite{ditzler2015learning} where the data generating process is evolving over time. Formally, a drift corresponds to a change in the joint probability $p(x,y)$. Despite that drift can manifest itself in other forms, this work focuses on $p(y)$ drift (i.e. a change in the prior probability) because such a change can lead to class imbalance. Note that the \textit{true} decision boundary remains unaffected when $p(y)$ drift occurs, however, the classifier's \textit{learnt} boundary may drift away from the true one.

\subsection{Online Class Imbalance Learning}\label{sec:related_work}
The majority of existing work addresses class imbalance in offline learning, while some others require chunk-based data processing \cite{wang2018systematic,hoens2012learning}. Little work deals with class imbalance in online learning and this section discusses the state-of-the-art.

The authors in \cite{wang2014cost} propose the cost-sensitive online gradient descent ($CSODG$) method that uses the following loss function:
\begin{equation}
J = ( I_{y^t=0} + I_{y^t=1} \frac{w_p}{w_n} ) ~ l(y^t, \hat{y}^t)
\end{equation}

\noindent where $I_{condition}$ is the indicator function that returns 1 if $condition$ is satisfied and 0 otherwise, $0 \leq w_p, w_n \leq 1$ and $w_p + w_n = 1$ are the costs for positive and negative classes respectively. The authors use the perceptron classifier and stochastic gradient descent, and apply the cost-sensitive modification to the hinge loss function achieving excellent results. The downside of this method is that the costs need to be pre-defined, however, the extent of the class imbalance problem may not be known in advance. In addition, it cannot cope with concept drift as the pre-defined costs remain static. In \cite{ghazikhani2013recursive}, the authors introduce $RLSACP$ which is a cost-sensitive perceptron-based classifier with an adaptive cost strategy.

A time decayed class size metric is defined in \cite{wang2015resampling} where for each class $c_k$, its size $s_k$ is updated at each time step $t$ according to the following equation:

\begin{equation}
s^t_k = \theta s^{t-1}_k + I_{y^t = c_k} (1 - \theta)
\end{equation}

\noindent  where $0 < \theta < 1$ is a pre-defined time decay factor that gives less emphasis on older data. This metric is used to determine the imbalance rate at any given time. For instance, for a binary classification problem where the positive class constitutes the minority, the imbalance rate at any given time $t$ is given by $s^t_p / s^t_n$.

Oversampling-based online bagging ($OOB$) is an ensemble method that adjusts the learning bias from the majority to the minority class adaptively through resampling by utilising the time decayed class size metric \cite{wang2015resampling}. An undersampling version called $UOB$ had also been proposed but was demonstrated to be unstable. $OOB$ with 50 neural networks has been shown to have superior performance. To determine the effectiveness of resampling solely, the authors examine the special case where there exists only a single classifier denoted by $OOB_{sg}$. Compared against the aforementioned $RLSACP$ and others, $OOB_{sg}$ has been shown to outperform the rest in the majority of the cases, thus concluding that resampling is the main reason behind the effectiveness of the ensemble \cite{wang2015resampling}.

Another approach to address drift is the use of sliding windows \cite{hoens2012learning}. It can be viewed as adding a memory component to the online learner; given a window of size $W$, it keeps in the memory the most recent $W$ examples. Despite being able to address concept drift, it is difficult to determine a priori the window size as a larger window is better suited for a slow drift, while a smaller window is suitable for a rapid drift. More sophisticated algorithms have been proposed, such as, a window of adaptable size or the use of multiple windows of different size \cite{lazarescu2004using}. The drawback of this approach is that it cannot handle class imbalance.

\begin{figure}[t]
	\centering
	\includegraphics[scale=0.45]{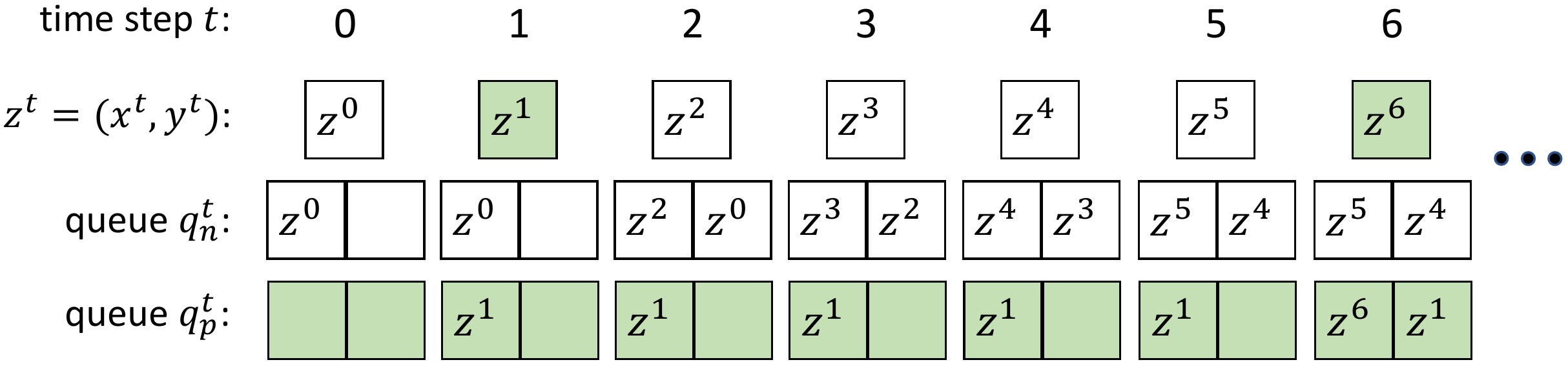}
	\caption{Example of $Queue_2$ resampling}
	\label{fig:q2_example}
\end{figure}

\section{Queue-based Resampling}
Online class imbalance learning is an emerging research topic and this work proposes queue-based resampling, a novel algorithm that addresses this problem. Focussing on binary classification, the central idea of the proposed resampling algorithm is to selectively include in the training set a subset of the positive and negative examples that appeared so far. Work closer to us is \cite{gao2008classifying} where the authors apply an analogous idea but in the context of chunk-based learning.

The selection of the examples is achieved by maintaining at any given time $t$ two separate queues of equal length $L \in \mathbb{Z}^+$, $q^t_n = \{(x_i,y_i)\}^L_{i=1}$ and $q^t_p = \{(x_i,y_i)\}^L_{i=1}$ that contain the negative and positive examples respectively. Let $z_i=(x_i,y_i)$, for any two $z_i, z_j \in q^t_n$ or ($q^t_p$) such that $j > i$, $z_j$ arrived more recently in time. Queue-based resampling stores the most recent example plus $2L-1$ old ones. We will refer to the proposed algorithm as $Queue_L$. Of particular interest is the special case $Queue_1$ where the length of each queue is $L=1$, as it has the major advantage of requiring just a single data point from the past.

An example demonstrating how $Queue_L$ works when $L=2$ is shown in Figure~\ref{fig:q2_example}. The upper part shows the examples that arrive at each time step e.g. $z^0$ and $z^6$ arrive at $t=0$ and $t=6$ respectively. Positive examples are shown in green. The bottom part shows the contents of each queue at each time step. Focussing on $t=5$, we can see that the queue $q^5_n$ contains the two most recent negative examples i.e. $z^4$ and $z^5$, and the queue $q^5_p$ contains the most recent positive example i.e. $z^1$ which is carried over since $t=1$.       

\begin{algorithm}[t]              
	\caption{Queue-based Resampling}   
	\label{alg:queue}                   
	\begin{algorithmic}[1]            
		
		\STATE{
			\textbf{Input:}\\
			maximum length $L$ of each queue\\
			queues $(q^t_p, q^t_n)$ for positive and negative examples
		}
		
		\FOR{each time step $t$}
		
		\STATE{receive example $x^t \in \mathbb{R}^d$}
		\STATE{predict class $\hat{y}^t \in \{0,1\}$}
		\STATE{receive true label $y^t \in \{0,1\}$}
		
		\STATE{let $z^t = (x^t,y^t)$}
		\IF{$y^t == 0$}
		\STATE{$q^t_n = q^{t-1}_n.append(z^t)$}
		\ELSE
		\STATE{$q^t_p = q^{t-1}_p.append(z^t)$}
		\ENDIF
		
		\STATE{let $q^t = q^t_p \cup q^t_n$ be the training set}
		\STATE{calculate cost on $q^t$ using Equation~\ref{eq:queue_cost}}
		\STATE{update classifier}
		
		\ENDFOR
		
	\end{algorithmic}
\end{algorithm}

The union of the two queues is then taken $q^t = q^t_p \cup q^t_n = \{(x_i,y_i)\}^{2L}_{i=1}$ to form the new training set for the classifier. The cost function is given in Equation~\ref{eq:queue_cost}:

\begin{equation}\label{eq:queue_cost}
J = \frac{1}{|q^t|} \sum_{i=1}^{|q^t|} l(y_i, h(x_i))
\end{equation}

\noindent where $|q^t| \leq 2L$ and $(x_i,y_i) \in q^t$.
At each time step the classifier is updated once according to the cost $J$ incurred i.e. a single update of the classifier's weights is performed. The pseudocode of our algorithm is shown in Algorithm~\ref{alg:queue}.

The effectiveness of queue-based resampling is attributed to a few important characteristics. Maintaining separate queues for each class helps to address the class imbalance problem. Including positive examples from the past in the most recent training set can be viewed as a form of oversampling. The fact that examples are propagated and carried over a series of time steps allows the classifier to `remember' old concepts. Additionally, to address the challenge of concept drift, the classifier needs to also be able to `forget' old concepts. This is achieved by bounding the length of queues to $L$, therefore, the queues are essentially behaving like sliding windows as well. Therefore, the proposed queue-based resampling method can cope with both class imbalance and concept drift.

\section{Experimental Setup}
Our experimental study\footnote{For reproducibility we publicly release our code and data: \url{https://github.com/kmalialis/queue_based_resampling} } is based on two popular synthetic datasets from the literature \cite{gama2004learning} where in both cases a classifier attempts to learn a non-linear decision boundary. These are, the Sine and Circle datasets and are described below.\\

\noindent \textbf{Sine.} It consists of two attributes $x$ and $y$ uniformly distributed in $[0,2\pi]$ and $[-1,1]$ respectively. The classification function is $y = sin(x)$. Instances below the curve are classified as positive and above the curve as negative. Feature rescaling has been performed so that $x$ and $y$ are in $[0, 1]$.\\

\noindent \textbf{Circle.} It has two attributes $x$ and $y$ that are uniformly distributed in $[0, 1]$. The circle function is given by $(x - x_c)^2 + (y - y_c)^2 = r_c^2$ where $(x_c, y_c)$ is its centre and $r_c$ its radius. The circle with $(x_c, y_c) = (0.4, 0.5)$ and $r_c = 0.2$ is created. Instances inside the circle are classified as positive and outside as negative.\\

Our baseline classifier is a neural network consisting of one hidden layer with eight neurons. Its configuration is as follows: $He$ \cite{he2015delving} weight initialisation, backpropagation and the $ADAM$ \cite{kingma2014adam} optimisation algorithms, learning rate of $0.01$, $Leaky ReLU$ \cite{maas2013rectifier} as the activation function of the hidden neurons, sigmoid activation for the output neuron, and the binary cross-entropy loss function.

\begin{table}[t]
	\centering
	\caption{Compared methods}\label{tab:methods}
	\begin{tabular}{|c|c|c|c|}
		\hline
		\textbf{Method} & \textbf{Class imbalance} & \textbf{Concept drift} & \textbf{Access to old data}\\
		\hline
		Baseline & no & no & no\\
		\hline
		Cost sensitive & yes & no & no\\
		\hline
		Sliding window & no & yes & yes ($W-1$)\\
		\hline
		$OOB_{sg}$ & yes & yes & no\\
		\hline\hline
		$Queue_1$ & yes & yes & yes (1)\\
		\hline
		$Queue_L$ & yes & yes & yes ($2L-1$)\\
		\hline
	\end{tabular}
\end{table}

For our study we implemented a series of state-of-the-art methods as described in Section~\ref{sec:related_work}. We implemented a cost sensitive version of the baseline which we will refer to as $CS$; the cost of the positive class is set to $\frac{w_p}{w_n}=\frac{0.95}{0.05}=19$ as in \cite{wang2014cost}. Furthermore, the sliding window method has been implemented with a window size of $W$. Moreover, the $OOB_{sg}$ has been implemented with the time decay factor set to $\theta = 0.99$ for calculating the class size at any given time.

For the proposed resampling method we will use the special case $Queue_1$ and another case $Queue_L$ where $L > 1$. Section~\ref{sec:queue_analysis} performs an analysis of $Queue_L$ by examining how the queue length $L$ affects the behaviour and performance of queue-based resampling. For a fair comparison with the sliding window method, we will set the window size to $W = 2L$ i.e. both methods will have access to the same amount of old data examples. A summary of the compared methods is shown in Table~\ref{tab:methods} indicating which methods are suitable for addressing class imbalance and concept drift. It also indicates whether methods require access to old data and, if yes, it includes the maximum number in the brackets.

A popular and suitable metric for evaluating algorithms under class imbalance is the geometric mean as it is not sensitive to the class distribution \cite{wang2018systematic}. It is defined as the geometric mean of recall and specificity. Recall is defined as the true positive rate ($R = \frac{TP}{P}$) and specificity is defined as the true negative rate ($S = \frac{TN}{N}$), where $TP$ and $P$ is the number of true positives and positives respectively, and similarly, $TN$ and $N$ for the true negatives and negatives. The geometric mean is then calculated using $G$-$mean = \sqrt{R \times S}$. To calculate the recall and specificity online, we use the prequential evaluation using fading factors as proposed in \cite{gama2013evaluating} and set the fading factor to $\alpha = 0.99$. In all graphs we plot the prequential $G$-$mean$ in every time step averaged over 30 runs, including the error bars showing the standard error around the mean.

\section{Experimental Results}

\begin{figure}[t]
	\centering
	
	\subfloat[$p(y=1)=0.5$]{\includegraphics[scale=0.38]{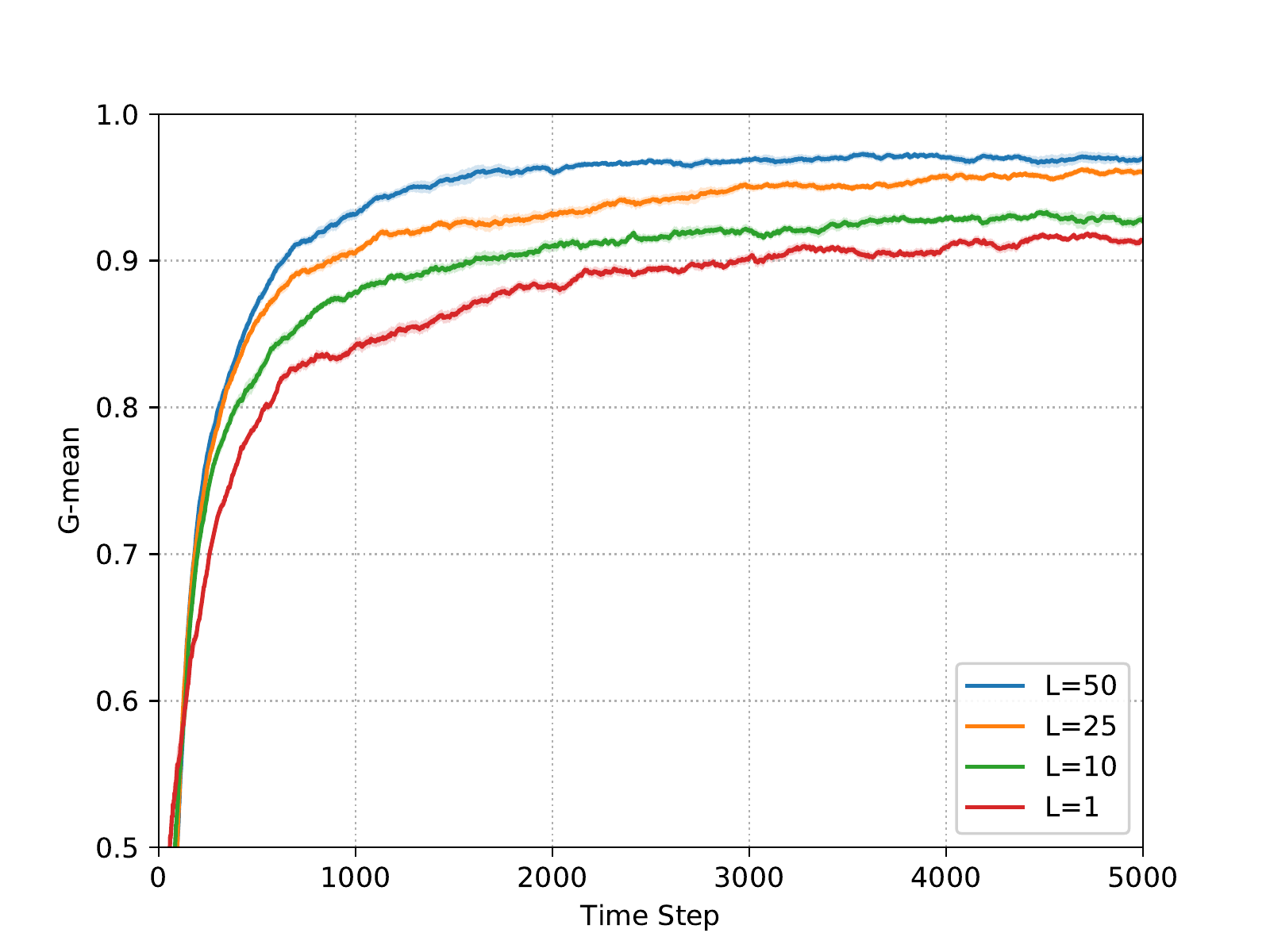}%
		\label{fig:queue_analysis_sine_50}}
	\subfloat[$p(y=1)=0.01$]{\includegraphics[scale=0.38]{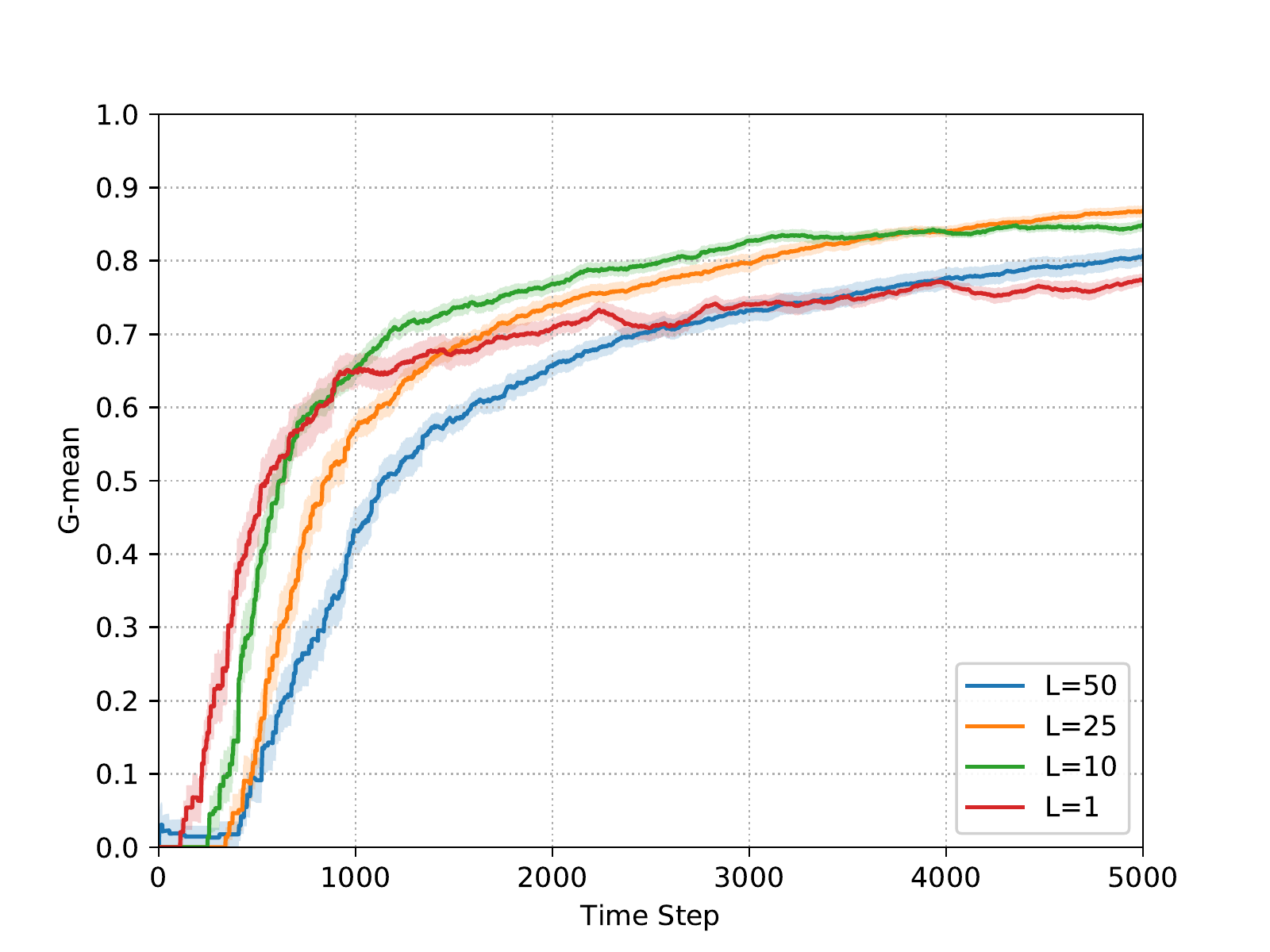}%
		\label{fig:queue_analysis_sine_1}}
	
	\caption{Effect of queue length on the Sine dataset}
\end{figure}

\subsection{Analysis of Queue-based Resampling}\label{sec:queue_analysis}
In this section we investigate the behaviour of $Queue_L$ resampling under various queue lengths ($L \in [1, 10, 25, 50]$) and examine how these affect its performance. Furthermore, we consider a balanced scenario (i.e. $p(y=1)=0.5$) and a scenario with a severe class imbalance of $1\%$ (i.e. $p(y=1)=0.01$).

Figures~\ref{fig:queue_analysis_sine_50} and~\ref{fig:queue_analysis_sine_1} depict the behaviour of the proposed method on the balanced and severely imbalanced scenario respectively for the Sine dataset. It can be observed from Figure~\ref{fig:queue_analysis_sine_50} that the larger the queue length the better the performance, specifically, the best performance is achieved when $L=50$. It can be observed from Figure~\ref{fig:queue_analysis_sine_1} that the smaller the queue length the faster the learning speed. $Queue_1$ dominates in the first 500 time steps, however, its end performance is inferior to the rest. The method with $L=10$ dominates for over 3000 steps. Given additional learning time the method with $L=25$ achieves the best performance. The method with $L=50$ is unable to outperform the one with $L=10$ after 5000 steps, in fact, it performs similarly to $Queue_1$.

It is important to emphasise that contrary to offline learning where the end performance is of particular concern, in online learning both the end performance and learning time are of high importance. For this reason, we have decided to focus on $Queue_{25}$ as it constitutes a reasonable trade-off between learning speed and performance. As already mentioned, we will also focus on $Queue_1$ as it has the advantage of requiring only one data example from the past.

\subsection{Comparative Study}
\begin{figure}[t]
	\centering
	
	\subfloat[$p(y=1)=0.1$]{\includegraphics[scale=0.38]{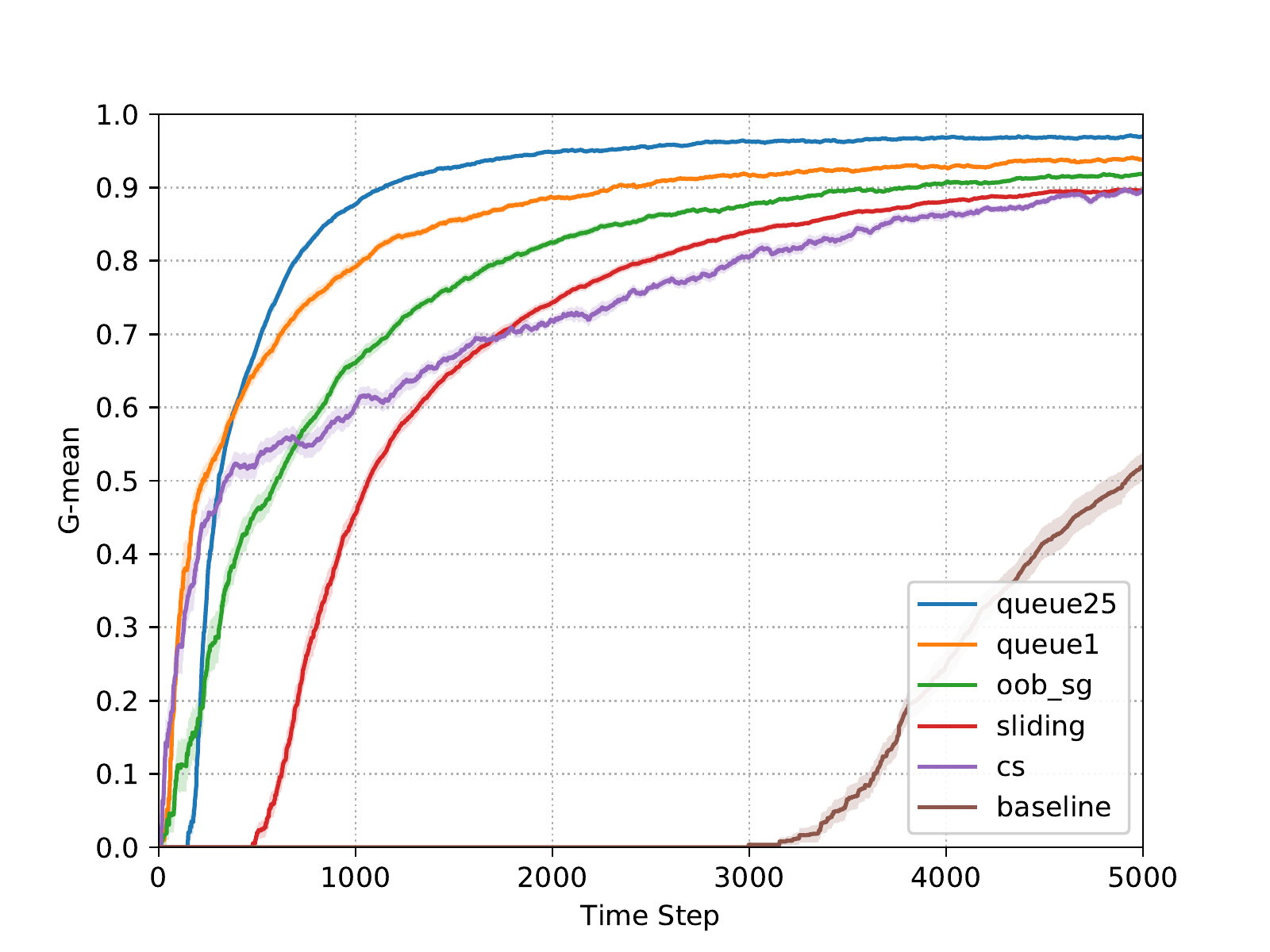}%
		\label{fig:comparison_circle_imbalance_10}}
	\subfloat[$p(y=1)=0.01$]{\includegraphics[scale=0.38]{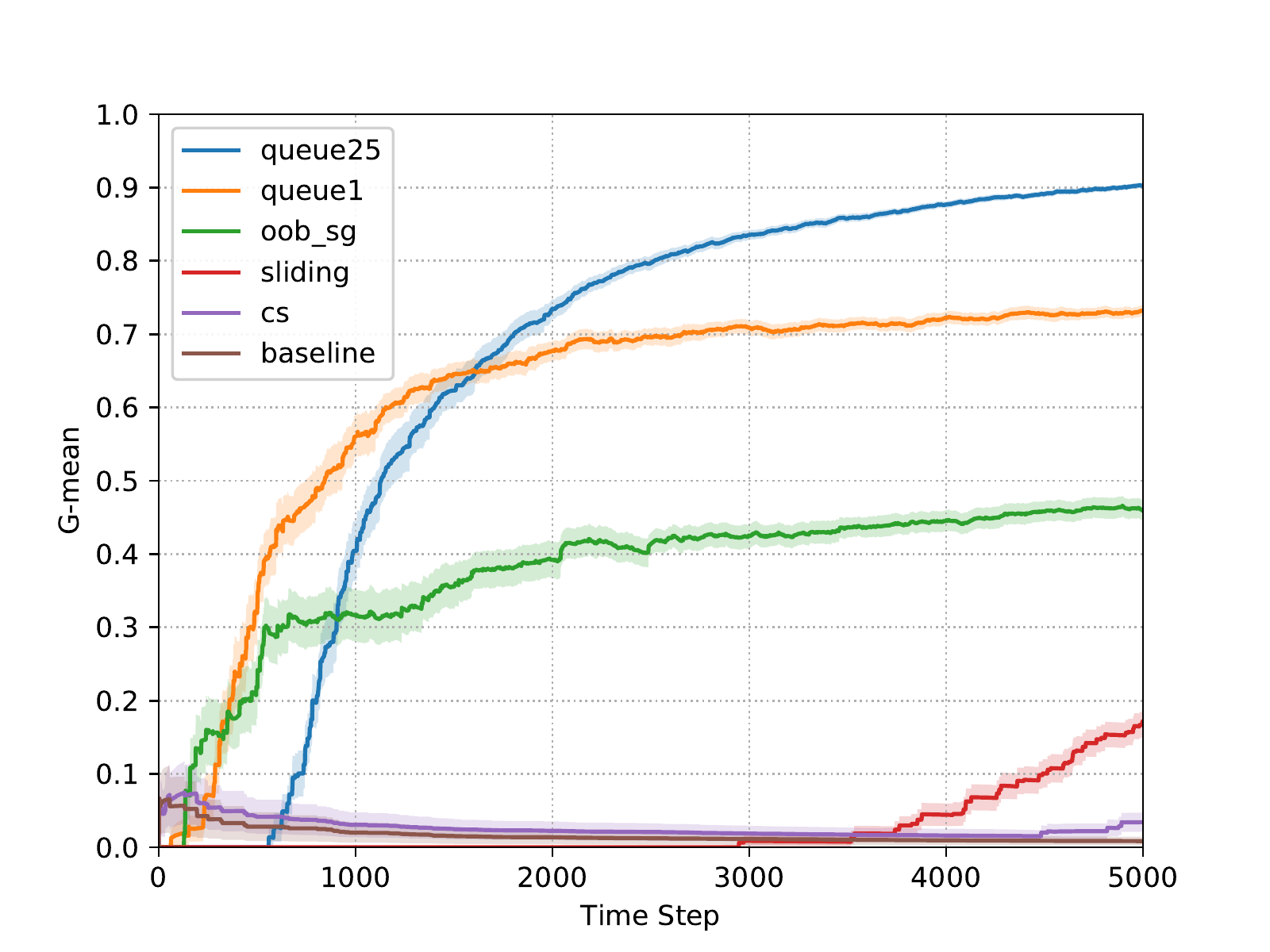}%
		\label{fig:comparison_circle_imbalance_1}}
	
	\caption{Class imbalance on the Circle dataset}
\end{figure}

Figure~\ref{fig:comparison_circle_imbalance_10} depicts a comparative study of all the methods in the scenario involving $10\%$ class imbalance for the Circle dataset. The baseline method, as expected, does not perform well and only starts learning after about 3000 time steps. The proposed $Queue_{25}$ has the best performance at the expense of a late start. $Queue_1$ also outperforms the rest although towards the end other methods like $OOB_{sg}$ close the gap. Similar results are obtained for the Sine dataset but are not presented here due to space constraints.

Figure~\ref{fig:comparison_circle_imbalance_1} shows how each method compares to each other in the $1\%$ class imbalance scenario. Both the proposed methods outperform the state-of-the-art $OOB_{sg}$. Despite the fact that $Queue_{25}$ performs considerably better than $Queue_1$, it requires about 1500 time steps to surpass it. Additionally, we stress out that $Queue_1$ only requires access to a single old example.

\begin{figure}[t]
	\centering
	
	\subfloat[Sine dataset]{\includegraphics[scale=0.38]{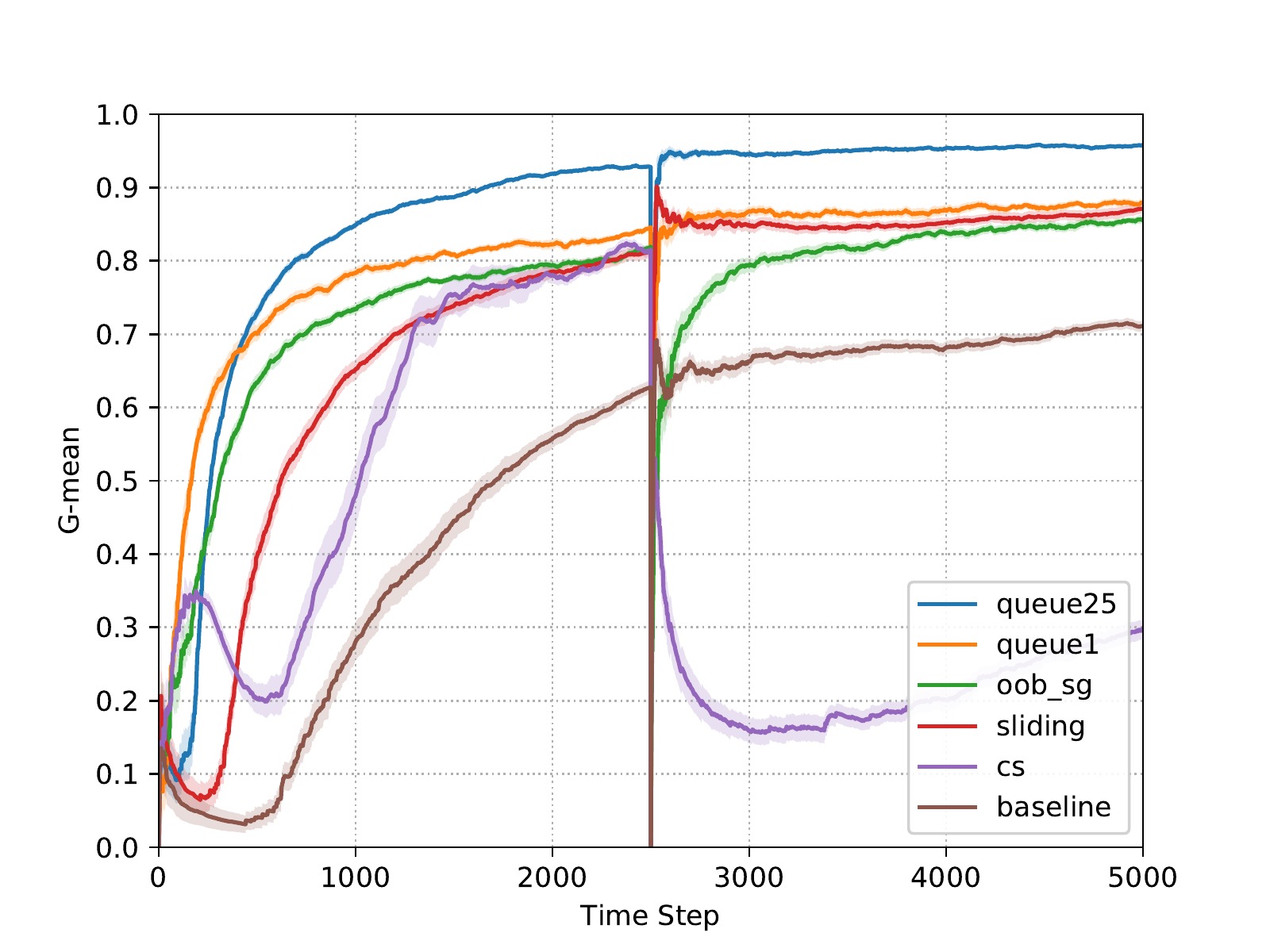}%
		\label{fig:comparison_sine_drift_abrupt_high}}
	\subfloat[Circle dataset]{\includegraphics[scale=0.38]{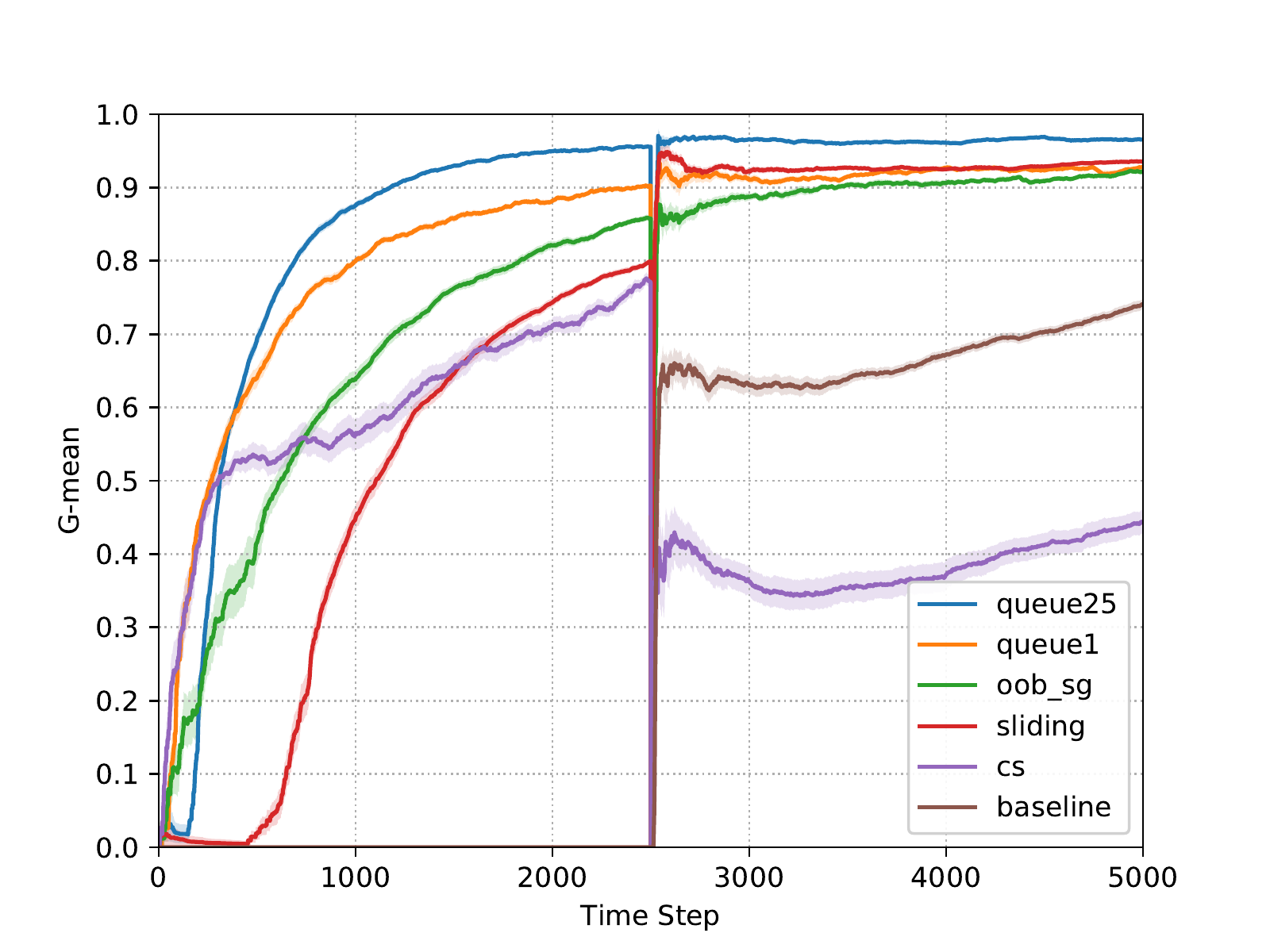}%
		\label{fig:comparison_circle_drift_abrupt_high}}
	
	\caption{Class imbalance and concept drift}
\end{figure}

We now examine the behaviour of all methods in the presence of both class imbalance and drift. Figures~\ref{fig:comparison_sine_drift_abrupt_high} and~\ref{fig:comparison_circle_drift_abrupt_high} show the performance of all methods for the Sine and Circle datasets respectively. Initially, class imbalance is $p(y=1) = 0.1$ but at time step $t=2500$ an abrupt drift occurs and this becomes $p(y=1) = 0.9$. At the time of drift we reset the prequential $G$-$mean$ to zero, thus ensuring the performance observed remains unaffected by the performance prior the drift \cite{wang2015resampling}. Similar results are observed for both datasets. $Queue_{25}$ outperforms the rest at the expense of a late start. $Queue_1$ starts learning fast, initially it outperforms other methods but their end performance is close. $OOB_{sg}$ is affected more by the drift in the Sine dataset but recovers soon. The baseline method outperforms its cost sensitive version after the drift because the pre-defined costs of method $CS$ are no longer suitable in the new situation.

\section{Conclusion}
Online class imbalance learning constitutes a new problem and an emerging research topic. We propose a novel algorithm, queue-based resamping, to address this problem. Focussing on binary classification problems, the central idea behind queue-based resampling is to selectively include in the training set a subset of the negative and positive examples by maintaining at any given time a separate queue for each class. It has been shown to outperform state-of-the-art methods, particularly, in scenarios with severe class imbalance. It has also been demonstrated to work well when abrupt concept drift occurs. Future work will examine the behaviour of queue-based resampling in various other types of concept drift (e.g. gradual). A challenge faced in the area of monitoring of critical infrastructures is that the true label of examples can be noisy or even not available. We plan to address this challenge in the future.

\subsubsection*{Acknowledgements}
This work has been supported by the European Union's Horizon 2020 research and innovation programme under grant agreement No 739551 (KIOS CoE) and from the Republic of Cyprus through the Directorate General for European Programmes, Coordination and Development.

%
%
%

\bibliographystyle{splncs04}
\bibliography{mybib}

\begin{thebibliography}{10}
\providecommand{\url}[1]{\texttt{#1}}
\providecommand{\urlprefix}{URL }
\providecommand{\doi}[1]{https://doi.org/#1}

\bibitem{ditzler2015learning}
Ditzler, G., Roveri, M., Alippi, C., Polikar, R.: Learning in nonstationary
  environments: A survey. IEEE Computational Intelligence Magazine
  \textbf{10}(4),  12--25 (2015)

\bibitem{gama2004learning}
Gama, J., Medas, P., Castillo, G., Rodrigues, P.: Learning with drift
  detection. In: Brazilian symposium on artificial intelligence. pp. 286--295.
  Springer (2004)

\bibitem{gama2013evaluating}
Gama, J., Sebasti{\~a}o, R., Rodrigues, P.P.: On evaluating stream learning
  algorithms. Machine learning  \textbf{90}(3),  317--346 (2013)

\bibitem{gao2008classifying}
Gao, J., Ding, B., Fan, W., Han, J., Philip, S.Y.: Classifying data streams
  with skewed class distributions and concept drifts. IEEE Internet Computing
  \textbf{12}(6) (2008)

\bibitem{ghazikhani2013recursive}
Ghazikhani, A., Monsefi, R., Yazdi, H.S.: Recursive least square perceptron
  model for non-stationary and imbalanced data stream classification. Evolving
  Systems  \textbf{4}(2),  119--131 (2013)

\bibitem{he2008learning}
He, H., Garcia, E.A.: Learning from imbalanced data. IEEE Transactions on
  Knowledge \& Data Engineering (9),  1263--1284 (2008)

\bibitem{he2015delving}
He, K., Zhang, X., Ren, S., Sun, J.: Delving deep into rectifiers: Surpassing
  human-level performance on imagenet classification. In: Proceedings of the
  IEEE international conference on computer vision. pp. 1026--1034 (2015)

\bibitem{hoens2012learning}
Hoens, T.R., Polikar, R., Chawla, N.V.: Learning from streaming data with
  concept drift and imbalance: an overview. Progress in Artificial Intelligence
   \textbf{1}(1),  89--101 (2012)

\bibitem{kingma2014adam}
Kingma, D.P., Ba, J.: Adam: A method for stochastic optimization. arXiv
  preprint arXiv:1412.6980  (2014)

\bibitem{kyriakides2014intelligent}
Kyriakides, E., Polycarpou, M.: Intelligent monitoring, control, and security
  of critical infrastructure systems, vol.~565. Springer (2014)

\bibitem{lazarescu2004using}
Lazarescu, M.M., Venkatesh, S., Bui, H.H.: Using multiple windows to track
  concept drift. Intelligent data analysis  \textbf{8}(1),  29--59 (2004)

\bibitem{maas2013rectifier}
Maas, A.L., Hannun, A.Y., Ng, A.Y.: Rectifier nonlinearities improve neural
  network acoustic models. In: Proc. icml. vol.~30, p.~3 (2013)

\bibitem{mao2015online}
Mao, W., Wang, J., Wang, L.: Online sequential classification of imbalanced
  data by combining extreme learning machine and improved smote algorithm. In:
  Neural Networks (IJCNN), 2015 International Joint Conference on. pp.~1--8.
  IEEE (2015)

\bibitem{wang2014cost}
Wang, J., Zhao, P., Hoi, S.C.: Cost-sensitive online classification. IEEE
  Transactions on Knowledge and Data Engineering  \textbf{26}(10),  2425--2438
  (2014)

\bibitem{wang2015resampling}
Wang, S., Minku, L.L., Yao, X.: Resampling-based ensemble methods for online
  class imbalance learning. IEEE Transactions on Knowledge and Data Engineering
   \textbf{27}(5),  1356--1368 (2015)

\bibitem{wang2018systematic}
Wang, S., Minku, L.L., Yao, X.: A systematic study of online class imbalance
  learning with concept drift. IEEE Transactions on Neural Networks and
  Learning Systems  (2018)

\bibitem{zhou2006training}
Zhou, Z.H., Liu, X.Y.: Training cost-sensitive neural networks with methods
  addressing the class imbalance problem. IEEE Transactions on Knowledge and
  Data Engineering  \textbf{18}(1),  63--77 (2006)

\end{thebibliography}

\end{document}